\documentclass[runningheads]{llncs}
\usepackage[T1]{fontenc}
\usepackage{graphicx}
\usepackage{hyperref}

\usepackage{todonotes}
\begin{document}
%
\title{Not Everything That Counts Can Be Counted: A~Case for Safe Qualitative AI}

\titlerunning{A Case for Safe Qualitative AI}
%
\author{Stine L. Beltoft\orcidID{0009-0006-5412-0050} \\ \and
Lukas Galke\orcidID{0000-0001-6124-1092}}

\authorrunning{S. Beltoft and L. Galke}
%
\institute{University of Southern Denmark\\
 \email{\{stinelb,galke\}@imada.sdu.dk}}
%
\maketitle              

\begin{abstract}
Artificial intelligence (AI) and large language models (LLM) are reshaping science, with most recent advances culminating in fully-automated scientific discovery pipelines. But qualitative research has been left behind. 
Researchers in qualitative methods are hesitant about AI adoption. Yet when they are willing to use AI at all, they have little choice but to rely on general-purpose tools like ChatGPT to assist with interview interpretation, data annotation, and topic modeling -- while simultaneously acknowledging these system's well-known limitations of being biased, opaque, irreproducible, and privacy-compromising.
This creates a critical gap: while AI has substantially advanced quantitative methods, the qualitative dimensions essential for meaning-making and comprehensive scientific understanding remain poorly integrated. We argue for developing dedicated qualitative AI systems built from the ground up for interpretive research. Such systems must be transparent, reproducible, and privacy-friendly. We review recent literature to show how existing automated discovery pipelines could be enhanced by robust qualitative capabilities, and identify key opportunities where safe qualitative AI could advance multidisciplinary and mixed-methods research.

\keywords{AI for Science  \and Mixed methods \and Large language models}
\end{abstract}
\section{Introduction}
In the past five years, artificial intelligence has made an undeniable impact on the practice of scientific research. The discourse surrounding AI in research has grown rapidly with students using ChatGPT to draft essays, scholars debating bias in large language models, and new guidelines on responsible AI for writing papers.
Much of this development, however, has focused on quantitative methods. AI is routinely used to analyze medical images, classify tumors, run complex statistical models, wrangle large numerical datasets, and gather literature for surveys -- everything from suggesting statistically sound question formats to simulating response patterns.
Most recently, these efforts have culminated in fully-automated scientific discovery pipelines~\cite{yamada2025ai,gottweis2025towards,lu_ai_2024}.
But even these pipelines remain fundamentally incomplete, as they lack a qualitative component.

Unlike quantitative research, where AI tools have become increasingly sophisticated and specialized, qualitative research currently relies on general-purpose AI tools like ChatGPT which were not built for the demands of interpretive, contextual inquiry and come with severe limitations regarding reproducibility (models are not deterministic and may become unavailable), transparency (bias-prone black box decisions), and privacy (cloud-based processing of sensitive data). This mismatch is one of the reasons why AI use in qualitative research often raises skepticism. Moreover, available tools do not support the epistemological depth, reflexivity, or human-centered reasoning at the heart of qualitative inquiry.

This current state, however, reflects design limitations rather than inherent AI constraints. AI has far greater potential for qualitative research, if developed with safety (transparency, reproducibility, privacy) and interpretive goals in mind. Ideally, such a tool would identify recurring themes across dozens of interviews and compare qualitative narratives to quantitative trends or even assist in the ethical and context-sensitive collection of qualitative data. Such tools could fundamentally expand, not replace, the capacities of qualitative researchers.

Here, we argue that \textit{if AI is to truly support scientific inquiry in its full breadth, it must do more than accelerate data processing. It must also engage with meaning, nuance, and complexity}. That requires understanding qualitative inquiry as requiring fundamentally different computational approaches and designing tools that serve its purposes from the ground up. After establishing the importance of qualitative research in Section~\ref{sec:why}, we structure our argument around the following key observations:

\begin{itemize}
\item Current practice: Researchers already use AI tools despite their limitations (Section~\ref{sec:qualitativetoday}) 
\item Contradictions in practice: AI seems to be used in qualitative research out of necessity, but simultaneously labeled as a limitation (Section~\ref{sec:contradictions})
\item Structural inequalities: The most powerful AI systems are biased towards high-resource languages such as English and aligned with Western norms and values (Section~\ref{sec:structuralinequalities}) 
\item Dangers of replacement: Synthetic participants undermine scientific validity (Section~\ref{sec:dangersofreplacement}) 
\item Quantification bias: AI for science and recent advances in automated scientific discovery pipelines exclude qualitative dimensions (Section~\ref{sec:quantbias})
\end{itemize}
\noindent
After these considerations, we give concrete examples, for how AI could in fact support and, thereby, advance the qualitative research process (Section~\ref{sec:designing}), before we conclude.

\section{Why qualitative research matters}\label{sec:why}
Qualitative research plays a vital role in scientific inquiry by offering tools and methods to explore meaning, context, and complexity, elements that often elude numerical representation. At its core, qualitative research is concerned with thick description, contextual meaning, lived experience, and interpretive reasoning\cite{geertz_thick_1973,denzin_sage_2009}. Rather than isolating variables or seeking predictive patterns, qualitative research attends to how people understand and construct their realities, often through narrative, contradiction, and ambiguity. 

Not everything of value in research can be counted or measured. Many of the most pressing questions in contemporary science like those related to mental health, social change, identity, and political opinion, require approaches that embrace nuance and plurality. In these domains, meaning often emerges not from aggregate data, but from situated knowledge and subjective accounts \cite{haraway_situated_1988,geertz_thick_1973}.

Qualitative research is considered both constructivist and constructionist~\cite{marshall_ethics_2024}: A constructivist approach focuses on how individuals create meaning based on their lived experiences \cite{bochner_unfurling_2018}, while a constructionist approach emphasizes how these meanings are shaped by broader cultural and social contexts \cite{lee_reconsidering_2012}. These epistemological commitments stand in contrast to positivist assumptions: qualitative research does not aim to discover a single, objective truth, but instead seeks to produce transferable insights that are situated, interpretive, and often provisional \cite{maxwell_qualitative_2013}. 

The qualitative research process typically involves stages such as data collection (e.g., interviews, observations), transcription, and interpretive analysis \cite{brinkmann_kvalitative_2010}. Each of these stages requires paying close attention to language, context, and power, and often involves iterative reflection by the researcher. This complexity and reflexivity are not flaws, they are essential features of a methodology designed to engage with human meaning in all its depth.

Despite its importance, qualitative research is often under-supported in computational settings. As this paper will argue, AI tools developed for scientific research, which are currently culminating in fully-automated scientific discovery pipelines~\cite{gottweis2025towards,lu_ai_2024,yamada2025ai}, must begin to take seriously the methodological and epistemic demands of qualitative inquiry, rather than treating it as an afterthought or reducing it to mere text processing.

\section{AI in the qualitative research pipeline today}\label{sec:qualitativetoday}
A small-but steadily growing set of systems is being built specifically for the interpretive tasks that define qualitative research (described in more detail below). Although most remain at prototype stage or are add-ons to established software, they illustrate what an explicitly qualitative AI ecosystem could look like.

Interviewbot~\cite{beltoft_interview_2025} shows how a large language model can be tuned for semi-structured interviewing. The system generates follow-up questions in real time, aiming to preserve conversational flow while leaving the researcher in control of topic depth and ethical boundaries.
Cody \cite{rietz_cody_2021} tackles one of the most labour-intensive steps in qualitative analysis; coding. It combines rule-based logic with supervised machine learning to propose initial codes, surface disagreements, and log provenance so that researchers can audit every suggestion before acceptance. 

Moving into observations, The Ethnobot \cite{tallyn_ethnobot_2018} embeds a chatbot in Internet-of-Things devices to capture contextual anecdotes from users as they interact with smart-mobility services. By prompting short, situated reflections, it augments traditional field-notes with time-stamped micro-ethnographies.
JourneyBot \cite{hwang_journeybot_2023} demonstrates AI-assisted sense-making. The tool lets design-research teams feed interview excerpts into an interactive journey-map visualizer; a transformer model highlights convergent pain points and alternative storylines without flattening contradictory accounts.

Two mainstream qualitative analysis suites are beginning to natively integrate AI. MAXQDA’s \cite{verbi_maxqda_nodate} experimental ``AI Assist'' module uses large language models to suggest code–text matches and memo prompts, while NVivo AI \cite{qsr_international_nvivo_nodate} offers auto-summaries of interviews in multiple languages and proposes sentiment categories that analysts can accept, revise, or delete. Because both platforms store data locally (or on researcher-controlled servers), they can be configured for GDPR-compliant projects, an essential step for sensitive fieldwork. \newline

Taken together, these examples signal a shift from re-purposing general NLP tools toward building software that recognizes qualitative epistemologies: context-sensitive questioning, researcher reflexivity, and transparent human-AI collaboration. Yet they also reveal gaps. Most current systems, operate on English-language corpora and struggle with multilingual or oral data, excel at first-pass coding or prompting but rarely support deeper theoretical integration, and remain locked inside proprietary ecosystems, limiting community scrutiny and method sharing.
Bridging these gaps will require dedicated efforts, open-source benchmarks grounded in qualitative theory, and closer partnership between computer scientists and interpretivist researchers. Only then, AI can become a reliable partner in the nuanced work of qualitative inquiry.

\section{Contradictions in practice: The ethics and use of AI in qualitative research}\label{sec:contradictions}
Despite growing interest in AI across disciplines, AI for qualitative research is deeply contested. Researchers routinely raise concerns about its use with fears of epistemic reductionism, replication of bias, and the erosion of interpretive integrity. These critiques appear across disciplines, often centered on the limitations of opaque large language models like ChatGPT~\cite{friedman2024should,morgan2023exploring}. Besides criticism regarding bias of the models, discussed in detail in Section~\ref{sec:structuralinequalities}, the use of such opaque models comes with concerns around the following three main aspects: 
\begin{enumerate}
    \item \textbf{Transparency.} The models are not openly available, their training data is unknown, and it is not clear to what extent web interfaces employ internal routing (e.g., relaying supposedly easy queries to smaller and weaker models).
    \item \textbf{Reproducibility.} As these AI systems can simply cease being available (e.g., models being replaced by a newer version), it is almost impossible to ensure reproducibility when using commercial general-purpose AI systems.
    \item \textbf{Privacy.} For the most powerful AI systems, requests need to be sent to central compute clusters, and could leak into the training data of the models. While agreements can be made, sensitive data is still most often processed in areas where different privacy regulations apply.
\end{enumerate}
Indeed, papers on the ethical implications of generative AI in qualitative research have proliferated, with many authors arguing that such tools should be used cautiously, if at all~\cite{marshall_ethics_2024}.

Yet these ethical warnings have not prevented the widespread and growing use of these tools in practice. \textit{Many qualitative researchers now incorporate generative AI into their workflows, using it for tasks such as interview summarization, preliminary coding, literature synthesis, and manuscript drafting \cite{lee_harnessing_2024,naeem_thematic_2025,turobov_using_2024,xiao_supporting_2023}.} Paradoxically, these uses are often acknowledged in limitations sections only to be disavowed moments later, framed as experimental, provisional, or morally suspect. The result is a kind of institutionalized ambivalence: AI is used out of necessity or convenience, while its legitimacy is publicly questioned. This tension points to a deeper structural contradiction in current research culture. 

The gap between public disapproval and private adoption reveals a failure not of ethics, but of infrastructure. Researchers are turning to AI because the tools that would truly support qualitative inquiry do not yet exist. The only available systems, such as ChatGPT, were not designed for qualitative science, and their use reflects a lack of better alternatives rather than a wholehearted endorsement.

Meanwhile, industry sectors that could substantially benefit from qualitative AI, such as healthcare, UX design, education, and policy, remain hesitant to adopt these tools at scale. This hesitancy is not due to a lack of need or potential, but rather due to economic, reputational, and legal concerns. \textit{There is uncertainty around data privacy, fear of backlash for using AI in domains tied to empathy or ethics, and a general lack of confidence in current tools’ validity}. As a result, adoption remains cautious and fragmented, even in spaces where AI could augment the work of human-centered analysis.

The situation is further compounded by the lack of investment in building qualitative AI tools within computer science itself. The stigma around qualitative reasoning as subjective, unscalable, or unscientific has discouraged many technical researchers from engaging seriously with the epistemic goals of interpretive inquiry. While quantitative AI enjoys decades of methodological refinement and dedicated tooling, qualitative AI is often dismissed as mere “text generation” and left to general-purpose models that are ill-equipped for the nuances of fieldwork, narrative, and theory.

This has left the field in a cycle of critique without construction. Scholars highlight the risks of qualitative AI while quietly relying on it; developers avoid building qualitative tools due to perceived reputational risk or unclear metrics of success. The result is a landscape where qualitative research is both underserved and overexposed, used widely, defended rarely, and rarely built for from the ground up. To move forward, we must break this cycle by shifting the conversation from whether qualitative researchers should use AI to how we can design AI systems that actually reflect the interpretive, reflexive, and situated nature of qualitative research.

\section{Structural inequalities and epistemic loss}\label{sec:structuralinequalities}
The increasing use of AI in qualitative research raises not only methodological concerns, but also critical questions about whose knowledge is being represented, and whose is being erased~\cite{barikeri2021redditbias,gallegos2024bias} (inter alia). When lived experiences are replaced by model-generated outputs, there is a real risk that the perspectives of marginalized, underrepresented, or politically contested groups are lost in translation. The apparent neutrality and fluency of AI-generated text can mask a deeper structural problem: the reproduction of dominant narratives at the expense of dissenting or non-normative voices.

Large language models like ChatGPT are trained on enormous corpora of text from the internet, academic literature, news, and social media~\cite{raffel2020exploring}. While this suggests some kind of democratic inclusivity, in practice it mirrors and amplifies the imbalances already present in those sources~\cite{bender2021dangers}. Groups with less access to publication channels, fewer mentions in data-rich environments, or voices that exist primarily in oral, localized, or situated contexts are less likely to be represented, or are represented inaccurately. This leads to an epistemic loss that privileges what is already visible, already legible, and already powerful.

AI systems that are not grounded in qualitative reasoning frameworks risk stripping away exactly the kind of positionality, contradiction, and contextual awareness that make qualitative research valuable in the first place. Without tools that account for situated knowledge and interpretive depth, we risk creating outputs that appear comprehensive but are, in fact, hollow, flattened representations of reality that obscure power dynamics rather than exposing them.

Trained on a vast body of human knowledge, technologies like ChatGPT are not a self that contains multitudes, but multitudes absent of a self~\cite{gillen2024}. By design, these tools cannot hold a single, describable point-of-view, and thus cannot offer the positionality necessary for trust, accountability, or critical engagement. In qualitative research, perspective is not a bug, it is a method. It is precisely through the recognition of voice, standpoints, and relational knowledge that qualitative inquiry makes power visible and subject to analysis.

When AI is used uncritically in place of human narratives, there is not only a risk of misrepresentation but also of epistemic loss: the gradual erasure of lived realities, local knowledge, and subaltern perspectives. Without intentional intervention, through the design of tools that preserve rather than overwrite human complexity, qualitative AI threatens to become a machine for amplifying existing inequalities, all while appearing neutral.

In this light, the development of qualitative AI is not merely a technical challenge, it is a political and ethical imperative. To support just and inclusive research, we must design AI systems that can recognize, preserve, and even foreground the contested, incomplete, and situated nature of knowledge itself.

\section{The dangers of replacing qualitative inquiry}\label{sec:dangersofreplacement}
As the capabilities of generative AI systems continue to grow, so does their use in domains once considered the exclusive domain of human interpretation, notably qualitative research. \textit{An emerging trend is the use of models like ChatGPT to simulate participant responses, particularly in exploratory studies where human data is costly, time-consuming, or ethically sensitive to collect}, as discussed in \cite{abdurahman2024perils,rossi2024problems}, for instance. In such cases, researchers have begun experimenting with AI as a proxy for human subjects, treating model-generated text as replacements for real-world opinions and perspectives.

This practice raises serious concerns: First, language models are not participants. They are not agents with lived experience, emotional investment, or sociocultural embeddedness. They generate plausible-sounding language by predicting patterns based on prior text data, but they cannot hold beliefs, be shaped by life events, or reflect on the world as social beings. This means that any insights they produce are simulations without grounding, textual echoes of discourse rather than situated narratives.
Take, for instance, a model trained on data up to 2020 asked to represent Danish public opinion toward the United States. It may claim that Danes ``generally have a favorable view of the U.S.'', reflecting sentiments prevalent at that time. However, this output ignores any subsequent political events, cultural shifts, or changes in public sentiment. The result is a temporally lagging, context-agnostic snapshot presented with the confidence of current truth, but based on outdated data and probabilistic assumptions.
This kind of simulation carries several dangers. First, there is the risk of misrepresentation: the views of real communities are overwritten by generic, statistically averaged language. Second, there is false confidence: because AI-generated responses are fluent and coherent, they may appear more authoritative than they are. Third, and most concerning, is the erasure of lived realities: When researchers substitute  actual participant voices for model outputs, they lose access to the richness, contradiction, and affect that define qualitative data.

The use of generative models as proxy for the human perspective fundamentally misunderstands what qualitative research is meant to do. Qualitative inquiry is not simply the collection of words: it is the practice of engaging with meaning in context, conversation, and relationship. It is attentive to temporality, voice, and positionality. These are dimensions that no current AI model can simulate with fidelity.

If we begin to treat models as equivalent to people, or model outputs as equivalent to data, we risk reducing qualitative inquiry to a kind of textual ventriloquism where synthetic language is mistaken for authentic experience. This is not only a methodological mistake but an ethical one, particularly in research contexts involving marginalized or vulnerable populations whose realities cannot and should not be approximated by statistical generalization.

Rather than replacing qualitative research, AI should be used to support it. The dangers of treating generative models as stand-ins for participants make clear that the future of AI in qualitative science depends not on mimicry, but on collaborative, grounded, and interpretive design. Only then can AI contribute meaningfully to the complexity that qualitative inquiry seeks to illuminate.

\section{Quantitative bias in AI for science}\label{sec:quantbias}
Artificial intelligence has become deeply integrated into the infrastructures of contemporary scientific discovery, but its development has been shaped by a strong, and often unacknowledged, quantitative bias. The most prominent AI tools used in science today are built to serve data-centric workflows. \textit{They perform tasks such as literature mining, hypothesis generation, statistical modeling, and simulation. From predicting protein structures \cite{jumper_highly_2021} to automating survey design \cite{jacobsen_chatbots_2025} and analyzing high-resolution images \cite{he_deep_2015}}, these tools are optimized for speed, scalability, and measurable outcomes. These systems are typically designed to support quantitative inference. Their strength lies in identifying patterns, classifying data, and making predictions from large, structured datasets. They excel at streamlining standardized processes: optimizing experimental design, calculating statistical correlations, and drawing generalizable conclusions from massive data volumes. \textit{This emphasis reflects long-standing assumptions within both AI and scientific research, that knowledge is best when it is objective, replicable, and quantifiable.} 

While this focus has undoubtedly advanced many fields, it has also created a significant asymmetry in how different forms of knowledge are supported. The vast majority of AI tools are designed for environments where variables are controlled, and data is clean. As a result, there is a striking lack of infrastructure for tasks fundamental to qualitative research: open-ended interpretation, coding of narrative data, theory-building, and contextual analysis. Qualitative research, which thrives on ambiguity, contradiction, and reflexivity, is often sidelined by systems that treat complexity as noise rather than insight.

This imbalance has broader epistemological consequences. When AI development in science disproportionately serves quantitative paradigms, it subtly reinforces the idea that interpretive or meaning-based inquiry is less scientific, less rigorous, or less important. Yet some of the most critical issues we face, how people experience injustice, how institutions shape behavior, how beliefs shift over time, are not well captured through statistical models alone. These are questions that require interpretive depth and contextual nuance. Moreover, this automation of quantitative reasoning often comes at the expense of interpretation. When large-scale pattern recognition replaces close reading, and when prediction supersedes understanding, we risk stripping research of its human-centered focus.

Interpretation is not a delay in the pipeline or an inefficiency to be optimized away, it is the space where meaning is constructed, where ethical questions are confronted, and where knowledge is made accountable to lived reality. Without AI tools designed to engage these interpretive processes, qualitative researchers are often forced to retrofit general-purpose systems, like generative language models, into workflows they were not built to support. These tools can be helpful in limited ways, transcribing interviews or summarizing texts, but they do not attend to the core epistemic practices of qualitative inquiry. The result is a patchwork approach that may satisfy surface-level efficiency but leaves the deeper work of meaning-making unsupported.

Recent work by recognizes similar limitations in current AI trajectories, proposing 'Scientist AI' systems that explain observations rather than merely act upon them~\cite{bengio2025superintelligent} -- an approach that aligns more closely with the interpretive goals of qualitative inquiry.

To move forward, the development of AI in science must recognize this imbalance and broaden its scope. A truly inclusive research pipeline would support not only the extraction of data, but also the construction of meaning. This requires an intentional investment in tools that respect and enable the complex, reflexive, and socially embedded nature of qualitative research.

\section{Designing AI to support qualitative research}\label{sec:designing}
If AI is to meaningfully contribute to the future of qualitative research, it must be designed not to replace interpretive processes, but to support, scaffold, and extend them. This requires a fundamental reorientation toward systems that are responsive to the specific epistemologies, ethics, and workflows of qualitative inquiry.
Rather than treating text as interchangeable data or analysis as reducible to classification, qualitative AI should engage with how meaning is produced, who it is produced by, and in what context. It must operate with the understanding that language is not only a medium of information, but also a medium of identity, ideology, and experience.
There is growing potential to build AI tools that are not just assistants, but thoughtful collaborators in the research process. For example interview coding assistants could help researchers surface recurring themes across transcripts while remaining transparent about ambiguity and disagreement in codes. Or theory-building collaborators that might suggest relationships between concepts or offer alternative framings drawn from specified theoretical traditions, rather than from general web-trained embeddings.
Even reflective tools that could prompt researchers to articulate their positionality, assumptions, or blind spots, making reflexivity a supported part of the analytic workflow rather than an afterthought. \\

Underlying these possibilities are a set of design principles that differ markedly from the assumptions built into most existing AI tools:
\begin{enumerate}
    \item \emph{Context-sensitivity} is key: systems should not treat language as decontextualized input but instead be aware of who is speaking, when, and in what setting.
    \item \emph{Temporal awareness} matters, especially in domains like social scientific or policy research, where attitudes and conditions change over time.
    \item All AI used in qualitative workflows should be \emph{human-in-the-loop} \cite{Mosqueira-Rey_Hernández-Pereira_Alonso-Ríos_Bobes-Bascarán_Fernández-Leal_2023}, ensuring that interpretation remains a human act, shaped by ethical judgment and domain expertise.
    \item Rather than present as another tool to master, AI should \emph{seamlessly extend} the self of the qualitative researcher \cite{schneider-Kamp_aisociety2025}.
    \item Qualitative AI must embrace \emph{non-reductive reasoning}: ambiguity, contradiction, and complexity are not bugs to be optimized away, but essential features of the data.
\end{enumerate}

These design principles must be underpinned by the technical foundations that current commercial systems lack: transparency, reproducibility, and privacy protection. Unlike the black-box ChatGPT queries that researchers currently resort to, qualitative AI systems should provide clear trails of their reasoning processes (i.e., explainability built-in), allowing researchers to understand and critique how interpretations were generated. They must further support reproducible workflows where the same data processed with the same parameters yields consistent results, which is impossible with constantly-updating, non-deterministic models. Moreover, these systems must be designed to run on local infrastructure, ensuring that sensitive interview data, personal narratives, and confidential fieldnotes never leave the researcher's control. This represents a fundamental departure from the cloud-based, privacy-compromising tools that researchers currently have no choice but to use when they turn to AI at all.

Going further, when qualitative AI systems are transparent and interoperable, they could finally enable the kind of interdisciplinary collaboration and mixed-methods integration that researchers have long wanted but struggled to achieve. Instead of forcing researchers to choose between quantitative precision and qualitative depth, properly designed systems could let them work seamlessly across both approaches. Imagine being able to trace how statistical patterns connect to narrative themes, with clear documentation of how each insight was generated and how they inform each other. This is a far cry from the current reality where quantitative and qualitative analyses typically happen in separate silos, often using completely different tools that have no interaction.

Rather than asking whether AI should be used in qualitative research, the more pressing question is: what kinds of AI are we willing to build? The future of qualitative AI depends on shifting from a culture of workaround to one of design, where tools are shaped in conversation with qualitative epistemologies, not in spite of them.

This vision aligns with recent calls for AI systems that prioritize understanding over action. Bengio et al.'s proposal for ``Scientist AI'' systems that explain the world from observations rather than simply generate text or take actions~\cite{bengio2025superintelligent} offers a compelling framework and anchor for designing safe qualitative AI . Such systems, designed to build explanatory models with explicit uncertainty quantification, naturally complement qualitative research's interpretive goals. Rather than replacing human judgment with automated decisions, these approaches could help researchers explore multiple theoretical framings, identify patterns across complex datasets, and articulate the limits of their interpretations, while maintaining the human researcher's central role in meaning-making.

\section{Conclusion}
Qualitative research remains an essential pillar of scientific understanding. It allows us to ask not only what is happening, but why, how, and for whom. It reveals the meaning behind behavior, the complexity behind consensus, and the voices often excluded from dominant narratives. As artificial intelligence continues to reshape research workflows, we must ensure that these technologies do not flatten science into what is merely countable, predictable, or extractable.

The current AI landscape reflects a lopsided investment in quantitative methods, an approach that, while powerful, is incomplete. To build a more inclusive and epistemically diverse research infrastructure, researchers, developers, and funders must turn their attention to qualitative AI: systems that support context, interpretation, reflexivity, and lived experience
while maintaining the transparency, reproducibility, and privacy protections that rigorous research demands. These tools are not simply technical add-ons but necessary complements to a scientific practice that aims to understand the world in its full human complexity.

Good science does not rely on a single way of knowing. It draws from numbers, narratives, models, and meanings. If AI is to serve science in its fullest sense, it must be built not only for efficiency, but also for empathy, ambiguity, and ethical depth. Supporting qualitative research through thoughtful AI design is not just a methodological improvement, it is a commitment to better, fairer, and more human-centered knowledge.

\subsubsection*{Acknowledgements}
This research has been in parts supported by the Danish Foundation Models project. Parts of the text (in particular, abstract and introduction) have been refined against critique from Claude Sonnet 4.
%
%
%
\bibliographystyle{splncs04}
\bibliography{references1,extra}

@inproceedings{barikeri2021redditbias,
  title={RedditBias: A Real-World Resource for Bias Evaluation and Debiasing of Conversational Language Models},
  author={Barikeri, Soumya and Lauscher, Anne and Vuli{\'c}, Ivan and Glava{\v{s}}, Goran},
  booktitle={Proceedings of the 59th Annual Meeting of the Association for Computational Linguistics and the 11th International Joint Conference on Natural Language Processing (Volume 1: Long Papers)},
  pages={1941--1955},
  year={2021}
}

@article{gallegos2024bias,
  title={Bias and fairness in large language models: A survey},
  author={Gallegos, Isabel O and Rossi, Ryan A and Barrow, Joe and Tanjim, Md Mehrab and Kim, Sungchul and Dernoncourt, Franck and Yu, Tong and Zhang, Ruiyi and Ahmed, Nesreen K},
  journal={Computational Linguistics},
  volume={50},
  number={3},
  pages={1097--1179},
  year={2024},
  publisher={MIT Press 255 Main Street, 9th Floor, Cambridge, Massachusetts 02142, USA~…}
}

@misc{gillen2024,
    author = {Gillen, Andrew L. Gillen},
    year = {2024},
    title ={Can We Trust AI in Qualitative Research?},
    url ={https://www.insidehighered.com/opinion/views/2024/10/09/can-we-trust-ai-qualitative-research-opinion},
    urldate = {2025-06-10},
    howpublished = {[Blog post]},
    publisher = {https://www.insidehighered.com/},
    note = {{I}nside Higher Education}
}

@article{schneider-Kamp_aisociety2025,
    title={The {AI}-extended professional self: user-centric {AI} integration into professional practice with exemplars from healthcare},
    ISSN={1435-5655},
    no_url={https://doi.org/10.1007/s00146-025-02319-5},
    DOI={10.1007/s00146-025-02319-5},
    journal={AI \& SOCIETY},
    author={Schneider-Kamp, Anna and Godono, Alessandro}, year={2025},
    month=mar,
    language={en}
}

@article{Mosqueira-Rey_Hernández-Pereira_Alonso-Ríos_Bobes-Bascarán_Fernández-Leal_2023,
  title={Human-in-the-loop machine learning: {A} state of the art},
  author={Mosqueira-Rey, Eduardo and Hern{\'a}ndez-Pereira, Elena and Alonso-R{\'\i}os, David and Bobes-Bascar{\'a}n, Jos{\'e} and Fern{\'a}ndez-Leal, {\'A}ngel},
  journal={Artificial Intelligence Review},
  volume={56},
  number={4},
  pages={3005--3054},
  year={2023},
  publisher={Springer}
}

@article{yamada2025ai,
  title={The {AI} {S}cientist-v2: {W}orkshop-level automated scientific discovery via agentic tree search},
  author={Yamada, Yutaro and Lange, Robert Tjarko and Lu, Cong and Hu, Shengran and Lu, Chris and Foerster, Jakob and Clune, Jeff and Ha, David},
  journal={arXiv preprint arXiv:2504.08066},
  year={2025}
}

@article{gottweis2025towards,
  title={Towards an {AI} co-scientist},
  author={Gottweis, Juraj and Weng, Wei-Hung and Daryin, Alexander and Tu, Tao and Palepu, Anil and Sirkovic, Petar and Myaskovsky, Artiom and Weissenberger, Felix and Rong, Keran and Tanno, Ryutaro and others},
  journal={arXiv preprint arXiv:2502.18864},
  year={2025}
}

@article{raffel2020exploring,
  title={Exploring the limits of transfer learning with a unified text-to-text transformer},
  author={Raffel, Colin and Shazeer, Noam and Roberts, Adam and Lee, Katherine and Narang, Sharan and Matena, Michael and Zhou, Yanqi and Li, Wei and Liu, Peter J},
  journal={Journal of machine learning research},
  volume={21},
  number={140},
  pages={1--67},
  year={2020}
}

@inproceedings{bender2021dangers,
  title={On the dangers of stochastic parrots: {C}an language models be too big?},
  author={Bender, Emily M and Gebru, Timnit and McMillan-Major, Angelina and Shmitchell, Shmargaret},
  booktitle={Proceedings of the 2021 ACM conference on fairness, accountability, and transparency},
  pages={610--623},
  year={2021}
}

@article{bengio2025superintelligent,
  title={Superintelligent agents pose catastrophic risks: {Can} {S}cientist {AI} offer a safer path?},
  author={Bengio, Yoshua and Cohen, Michael and Fornasiere, Damiano and Ghosn, Joumana and Greiner, Pietro and MacDermott, Matt and Mindermann, S{\"o}ren and Oberman, Adam and Richardson, Jesse and Richardson, Oliver and others},
  journal={arXiv preprint arXiv:2502.15657},
  year={2025}
}

@article{abdurahman2024perils,
  title={Perils and opportunities in using large language models in psychological research},
  author={Abdurahman, Suhaib and Atari, Mohammad and Karimi-Malekabadi, Farzan and Xue, Mona J and Trager, Jackson and Park, Peter S and Golazizian, Preni and Omrani, Ali and Dehghani, Morteza},
  journal={PNAS nexus},
  volume={3},
  number={7},
  pages={pgae245},
  year={2024},
  publisher={Oxford University Press US}
}

@article{rossi2024problems,
  title={The Problems of {LLM}-generated Data in Social Science Research},
  author={Rossi, Luca and Harrison, Katherine and Shklovski, Irina},
  journal={Sociologica},
  volume={18},
  number={2},
  pages={145--168},
  year={2024}
}

@article{friedman2024should,
  title={Should ChatGPT help with my research? A caution against artificial intelligence in qualitative analysis},
  author={Friedman, Carli and Owen, Aleksa and VanPuymbrouck, Laura},
  journal={Qualitative Research},
  no_pages={14687941241297375},
  year={2024},
  publisher={SAGE Publications Sage UK: London, England}
}

@article{morgan2023exploring,
  title={Exploring the use of artificial intelligence for qualitative data analysis: The case of ChatGPT},
  author={Morgan, David L},
  journal={International journal of qualitative methods},
  volume={22},
  no_pages={16094069231211248},
  year={2023},
  publisher={SAGE Publications Sage CA: Los Angeles, CA}
}

@book{denzin_sage_2009,
	address = {Thousand Oaks, Calif.},
	edition = {3. ed., [Nachdr.]},
	title = {The {Sage} handbook of qualitative research},
	isbn = {978-0-7619-2757-0},
	language = {eng},
	publisher = {Sage Publ},
	editor = {Denzin, Norman K. and Lincoln, Yvonna S. and {Sage Publications}},
	year = {2009},
}

@article{lee_harnessing_2024,
	title = {Harnessing {ChatGPT} for {Thematic} {Analysis}: {Are} {We} {Ready}?},
	volume = {26},
	issn = {1438-8871},
	shorttitle = {Harnessing {ChatGPT} for {Thematic} {Analysis}},
	url = {https://www.jmir.org/2024/1/e54974},
	doi = {10.2196/54974},
	language = {en},
	urldate = {2025-06-11},
	journal = {Journal of Medical Internet Research},
	author = {Lee, V Vien and Van Der Lubbe, Stephanie C C and Goh, Lay Hoon and Valderas, Jose Maria},
	month = may,
	year = {2024},
	pages = {e54974},
}

@article{naeem_thematic_2025,
	title = {Thematic {Analysis} and {Artificial} {Intelligence}: {A} {Step}-by-{Step} {Process} for {Using} {ChatGPT} in {Thematic} {Analysis}},
	volume = {24},
	issn = {1609-4069, 1609-4069},
	shorttitle = {Thematic {Analysis} and {Artificial} {Intelligence}},
	url = {https://journals.sagepub.com/doi/10.1177/16094069251333886},
	doi = {10.1177/16094069251333886},
	language = {en},
	urldate = {2025-06-11},
	journal = {International Journal of Qualitative Methods},
	author = {Naeem, Muhammad and Smith, Tracy and Thomas, Lorna},
	month = apr,
	year = {2025},
	pages = {16094069251333886},
}

@article{xiao_supporting_2023,
	title = {Supporting {Qualitative} {Analysis} with {Large} {Language} {Models}: {Combining} {Codebook} with {GPT}-3 for {Deductive} {Coding}},
	copyright = {Creative Commons Attribution 4.0 International},
	shorttitle = {Supporting {Qualitative} {Analysis} with {Large} {Language} {Models}},
	url = {https://arxiv.org/abs/2304.10548},
	doi = {10.48550/ARXIV.2304.10548},
	urldate = {2025-06-11},
	author = {Xiao, Ziang and Yuan, Xingdi and Liao, Q. Vera and Abdelghani, Rania and Oudeyer, Pierre-Yves},
	year = {2023},
	note = {Publisher: arXiv
Version Number: 1},
}

@misc{turobov_using_2024,
	title = {Using {ChatGPT} for {Thematic} {Analysis}},
	copyright = {Creative Commons Attribution 4.0 International},
	url = {https://arxiv.org/abs/2405.08828},
	doi = {10.48550/ARXIV.2405.08828},
	urldate = {2025-06-11},
	publisher = {arXiv},
	author = {Turobov, Aleksei and Coyle, Diane and Harding, Verity},
	year = {2024},
	note = {Version Number: 1},
}

@misc{he_deep_2015,
	title = {Deep {Residual} {Learning} for {Image} {Recognition}},
	copyright = {arXiv.org perpetual, non-exclusive license},
	url = {https://arxiv.org/abs/1512.03385},
	doi = {10.48550/ARXIV.1512.03385},
	urldate = {2025-06-10},
	publisher = {arXiv},
	author = {He, Kaiming and Zhang, Xiangyu and Ren, Shaoqing and Sun, Jian},
	year = {2015},
	note = {Version Number: 1},
}

@inproceedings{jacobsen_chatbots_2025,
	address = {Yokohama Japan},
	title = {Chatbots for {Data} {Collection} in {Surveys}: {A} {Comparison} of {Four} {Theory}-{Based} {Interview} {Probes}},
	isbn = {979-8-4007-1394-1},
	shorttitle = {Chatbots for {Data} {Collection} in {Surveys}},
	url = {https://dl.acm.org/doi/10.1145/3706598.3714128},
	doi = {10.1145/3706598.3714128},
	language = {en},
	urldate = {2025-06-10},
	booktitle = {Proceedings of the 2025 {CHI} {Conference} on {Human} {Factors} in {Computing} {Systems}},
	publisher = {ACM},
	author = {Jacobsen, Rune Møberg and Cox, Samuel Rhys and Griggio, Carla F. and Van Berkel, Niels},
	month = apr,
	year = {2025},
	pages = {1--21},
}

@article{jumper_highly_2021,
	title = {Highly accurate protein structure prediction with {AlphaFold}},
	volume = {596},
	issn = {0028-0836, 1476-4687},
	url = {https://www.nature.com/articles/s41586-021-03819-2},
	doi = {10.1038/s41586-021-03819-2},
	language = {en},
	number = {7873},
	urldate = {2025-06-10},
	journal = {Nature},
	author = {Jumper, John and Evans, Richard and Pritzel, Alexander and Green, Tim and Figurnov, Michael and Ronneberger, Olaf and Tunyasuvunakool, Kathryn and Bates, Russ and Žídek, Augustin and Potapenko, Anna and Bridgland, Alex and Meyer, Clemens and Kohl, Simon A. A. and Ballard, Andrew J. and Cowie, Andrew and Romera-Paredes, Bernardino and Nikolov, Stanislav and Jain, Rishub and Adler, Jonas and Back, Trevor and Petersen, Stig and Reiman, David and Clancy, Ellen and Zielinski, Michal and Steinegger, Martin and Pacholska, Michalina and Berghammer, Tamas and Bodenstein, Sebastian and Silver, David and Vinyals, Oriol and Senior, Andrew W. and Kavukcuoglu, Koray and Kohli, Pushmeet and Hassabis, Demis},
	month = aug,
	year = {2021},
	pages = {583--589},
}

@incollection{geertz_thick_1973,
	address = {New York},
	title = {Thick {Description}: {Towards} an {Interpretive} {Theory} of {Culture}},
	booktitle = {The {Interpretation} of {Cultures}},
	author = {Geertz, Clifford},
	year = {1973},
}

@article{haraway_situated_1988,
	title = {Situated {Knowledges}: {The} {Science} {Question} in {Feminism} and the {Privilege} of {Partial} {Perspective}},
	volume = {14},
	issn = {00463663},
	shorttitle = {Situated {Knowledges}},
	url = {https://www.jstor.org/stable/3178066?origin=crossref},
	doi = {10.2307/3178066},
	number = {3},
	urldate = {2025-06-10},
	journal = {Feminist Studies},
	author = {Haraway, Donna},
	year = {1988},
	pages = {575},
}

@misc{qsr_international_nvivo_nodate,
	title = {{NVivo}},
	url = {https://www.qsrinternational.com/nvivo-qualitative-data-analysis-software/home},
	publisher = {QSR International Pty Ltd},
	author = {QSR International},
}

@misc{verbi_maxqda_nodate,
	address = {Berlin, Germany},
	title = {{MAXQDA}},
	url = {https://www.maxqda.com/},
	publisher = {VERBI Software. Consult. Sozialforschung. GmbH},
	author = {VERBI},
}

@article{hwang_journeybot_2023,
	title = {{JourneyBot}: {Designing} a chatbot-driven interactive visualization tool for design research},
	copyright = {Creative Commons Attribution 4.0 International},
	shorttitle = {{JourneyBot}},
	url = {https://www.ijdesign.org/index.php/IJDesign/article/view/4674},
	doi = {10.57698/V17I3.06},
	urldate = {2025-05-21},
	author = {Hwang, Soojin and Kim, Dongwhan},
	collaborator = {Kim, Dongwhan},
	year = {2023},
	note = {Publisher: International Journal of Design},
}

@book{maxwell_qualitative_2013,
	edition = {3},
	title = {Qualitative research design: {An} interactive approach},
	publisher = {Sage},
	author = {Maxwell, J. A.},
	year = {2013},
}

@article{lee_reconsidering_2012,
	title = {Reconsidering {Constructivism} in {Qualitative} {Research}},
	volume = {44},
	issn = {0013-1857, 1469-5812},
	url = {https://www.tandfonline.com/doi/full/10.1111/j.1469-5812.2010.00720.x},
	doi = {10.1111/j.1469-5812.2010.00720.x},
	language = {en},
	number = {4},
	urldate = {2025-05-21},
	journal = {Educational Philosophy and Theory},
	author = {Lee, Cheu‐Jey George},
	month = jan,
	year = {2012},
	pages = {403--412},
}

@article{bochner_unfurling_2018,
	title = {Unfurling {Rigor}: {On} {Continuity} and {Change} in {Qualitative} {Inquiry}},
	volume = {24},
	issn = {1077-8004, 1552-7565},
	shorttitle = {Unfurling {Rigor}},
	url = {https://journals.sagepub.com/doi/10.1177/1077800417727766},
	doi = {10.1177/1077800417727766},
	language = {en},
	number = {6},
	urldate = {2025-05-21},
	journal = {Qualitative Inquiry},
	author = {Bochner, Arthur P.},
	month = jul,
	year = {2018},
	pages = {359--368},
}

@book{brinkmann_kvalitative_2010,
	address = {København},
	edition = {1},
	title = {Kvalitative {Metoder}: {En} {Grundbog}},
	publisher = {Hans Reitzels Forlag},
	author = {Brinkmann, Svend and Tanggaard, Lene},
	year = {2010},
}

@inproceedings{beltoft_interview_2025,
	address = {Porto, Portugal},
	title = {Interview {Bot}: {Can} {Agentic} {LLM}’s {Perform} {Ethnographic} {Interviews}?:},
	isbn = {978-989-758-737-5},
	shorttitle = {Interview {Bot}},
	url = {https://www.scitepress.org/DigitalLibrary/Link.aspx?doi=10.5220/0013387800003890},
	doi = {10.5220/0013387800003890},
	urldate = {2025-05-12},
	booktitle = {Proceedings of the 17th {International} {Conference} on {Agents} and {Artificial} {Intelligence}},
	publisher = {SCITEPRESS - Science and Technology Publications},
	author = {Beltoft, Stine and Schneider-Kamp, Peter and Askegaard, Søren},
	year = {2025},
	pages = {702--709},
}

@article{marshall_ethics_2024,
	title = {The {Ethics} of {Using} {Artificial} {Intelligence} in {Qualitative} {Research}},
	volume = {19},
	issn = {1556-2646, 1556-2654},
	url = {https://journals.sagepub.com/doi/10.1177/15562646241262659},
	doi = {10.1177/15562646241262659},
	language = {en},
	number = {3},
	urldate = {2025-05-07},
	journal = {Journal of Empirical Research on Human Research Ethics},
	author = {Marshall, David T. and Naff, David B.},
	month = jul,
	year = {2024},
	pages = {92--102},
}

@inproceedings{tallyn_ethnobot_2018,
	address = {Montreal QC Canada},
	title = {The {Ethnobot}: {Gathering} {Ethnographies} in the {Age} of {IoT}},
	isbn = {978-1-4503-5620-6},
	shorttitle = {The {Ethnobot}},
	url = {https://dl.acm.org/doi/10.1145/3173574.3174178},
	doi = {10.1145/3173574.3174178},
	language = {en},
	urldate = {2025-04-02},
	booktitle = {Proceedings of the 2018 {CHI} {Conference} on {Human} {Factors} in {Computing} {Systems}},
	publisher = {ACM},
	author = {Tallyn, Ella and Fried, Hector and Gianni, Rory and Isard, Amy and Speed, Chris},
	month = apr,
	year = {2018},
	pages = {1--13},
}

@misc{lu_ai_2024,
	title = {The {AI} {Scientist}: {Towards} {Fully} {Automated} {Open}-{Ended} {Scientific} {Discovery}},
	copyright = {Creative Commons Attribution 4.0 International},
	shorttitle = {The {AI} {Scientist}},
	url = {https://arxiv.org/abs/2408.06292},
	doi = {10.48550/ARXIV.2408.06292},
	urldate = {2025-04-02},
	publisher = {arXiv},
	author = {Lu, Chris and Lu, Cong and Lange, Robert Tjarko and Foerster, Jakob and Clune, Jeff and Ha, David},
	year = {2024},
	note = {Version Number: 3},
}

@inproceedings{rietz_cody_2021,
	address = {Yokohama Japan},
	title = {Cody: {An} {AI}-{Based} {System} to {Semi}-{Automate} {Coding} for {Qualitative} {Research}},
	isbn = {978-1-4503-8096-6},
	shorttitle = {Cody},
	url = {https://dl.acm.org/doi/10.1145/3411764.3445591},
	doi = {10.1145/3411764.3445591},
	language = {en},
	urldate = {2025-04-02},
	booktitle = {Proceedings of the 2021 {CHI} {Conference} on {Human} {Factors} in {Computing} {Systems}},
	publisher = {ACM},
	author = {Rietz, Tim and Maedche, Alexander},
	month = may,
	year = {2021},
	pages = {1--14},
}
%

\end{document}